\documentclass[runningheads]{llncs}
\usepackage{graphicx}
\usepackage{comment}
\usepackage{amsmath,amssymb} 
\usepackage{color}
\usepackage{subfigure}

\begin{document}
\pagestyle{headings}
\mainmatter
\def\ECCVSubNumber{4897}  

\title{Extended Batch Normalization} 

\titlerunning{Extended Batch Normalization}
%
\author{Chunjie Luo\inst{1,2} \and
Jianfeng Zhan\inst{1,2} \and
Lei Wang\inst{1} \and
Wanling Gao \inst{1}}
\authorrunning{Chunjie Luo et al.}
%
\institute{State Key Laboratory of Computer Architecture, \\
Institute of Computing Technology, Chinese Academy of Sciences
\email{\{luochunjie,zhanjianfeng,wanglei\_2011,gaowanling\}@ict.ac.cn} \and
University of Chinese Academy of Sciences }

%
%
\maketitle

\begin{abstract}
Batch normalization (BN) has become a standard technique for training the modern deep networks. However, its effectiveness diminishes when the batch size becomes smaller, since the batch statistics estimation becomes inaccurate. That hinders batch normalization’s usage for 1) training larger model which requires small batches constrained by memory consumption, 2) training on mobile or embedded devices of which the memory resource is limited. 
In this paper, we propose a simple but effective method, called extended batch normalization (EBN).  
For NCHW format feature maps, extended batch normalization computes the mean along the (N, H, W) dimensions, as the same as batch normalization, to maintain the advantage of batch normalization.  To alleviate the problem caused by small batch size, extended batch normalization computes the standard deviation along the (N, C, H, W) dimensions, thus enlarges the number of samples from which the standard deviation is computed. We compare extended batch normalization with batch normalization and group normalization on the datasets of MNIST, CIFAR-10/100, STL-10, and ImageNet, respectively. The experiments show that extended batch normalization alleviates the problem of batch normalization with small batch size while achieving close performances to batch normalization with large batch size.
\keywords{Neural networks, Normalization}
\end{abstract}

\section{Introduction}

Deep neural networks have received great success in many areas. Batch normalization (BN) has become a standard technique for training the modern deep networks. 
It normalizes the features by the mean and the standard deviation computed within a batch of samples. 
When training with batch normalization, examples are seen in conjunction with each other in the minibatch. The model looks at multiple training examples in integration, rather than in isolation. The coordination between examples helps the learning process. Moreover, the random selection of examples in the minibatch brings the sampling noises, that provides similar regularization benefits as Dropout \cite{srivastava2014dropout}.

However, its effectiveness diminishes when the batch size becomes smaller, since the noises are too much and make inaccurate batch statistics estimation. That hinders batch normalization’s usage for 
1) training larger (deeper and wider) models which requires small batches constrained by memory consumption. 
2) training on mobile or embedded devices of which the memory resource is limited.
Training model on edge devices has received more and more attention. For example, 
federated learning, a hot topic in machine learning,  aims at training a model across multiple decentralized edge devices or servers to address privacy and security issues. The heterogeneous environments require a robust training algorithm with large or small batch size.
 
Several normalization methods have been proposed to address the problem caused by small batch size, e.g. normalization propagation \cite{arpit2016normalization}, batch renormalization \cite{ioffe2017batch}, kalman normalization \cite{wang2018kalman}. The complexity limits their usage in the community. Group normalization \cite{wu2018group} is simple and effective method to handle the training with small batch size. However, there is a extra hyper parameter $G$ (the number of group) to be tuned in group normalization. When $G$ is large, the number of channels per group decreases. Then the estimation of the mean and the standard deviation (std) may be inaccurate. As $G$ equals to the number of channel of the layer, group normalization becomes instance normalization \cite{ulyanov2016instance}. When $G$ is small, the number of channels in the same group increases, but this makes the channels more correlative with each other. When $G$ equals to 1, group normalization becomes layer normalization \cite{ba2016layer}. 

In this paper, we propose a simple  but effective method, called extended batch normalization (EBN). The key difference between extended batch normalization and other normalization methods is that extended batch normalization compute the mean and the standard deviation from different set of pixels. 
For NCHW format feature, let N refer to batch dimension, C refer to channel dimension, H and W refer to the spatial height and width dimensions.
To maintain the advantage of batch normalization, extended batch normalization computes the mean along the (N,H,W) dimensions just as the same as batch normalization. To alleviate the problem caused by small batch size, extended batch normalization computes the standard deviation along the (N, C, H, W) dimensions, thus enlarges the number of pixel samples from which the standard deviation is computed.

At inference time for extended batch normalization, the mean and the standard deviation are pre-computed
from the training set by moving average as the same as batch normalization. There is no need to compute the mean and the standard deviation at inference time comparing to group normalization.  Moreover, since the mean and the standard deviation are pre-computed and fixed at inference time, the normalization can be fused into convolution operation. That is very helpful to speed up the inference especially on the mobile or embedded devices.

We compare extended batch normalization with batch normalization and group normalization on the datasets of MNIST, CIFAR-10/100, STL-10, and ImageNet, respectively. The experiments show that extended batch normalization alleviates the problem of batch normalization with small batch size while achieving close performances to batch normalization with large batch size.

\section{Related Work}

Batch normalization \cite{ioffe2015batch} is effective at accelerating and
improving the training of deep neural networks by reducing internal covariate shift. It performs the normalization for each training minibatch along (N,H,W) dimensions in the case of NCHW format feature. Since batch normalization uses the statistics on minibatch examples, its effect is dependent on the minibatch size. 

To mitigate this problem, normalization propagation \cite{arpit2016normalization} uses a data-independent parametric estimate of the mean and standard deviation instead of explicitly calculating from data. Normalization propagation depends on the strong assumptions that the activation values follow Gaussian distribution and the weight matrix of hidden layers are roughly incoherent. 

Batch renormalization \cite{ioffe2017batch} introduces two extra parameters to correct the fact that the minibatch statistics differ from the population ones.
It needs train the model for a certain number of iterations with batch normalization alone, without the correction, then ramp up the amount of allowed correction.

There is a family of methods which avoid normalizing along the batch dimension. Layer normalization \cite{ba2016layer} computes the mean and standard deviation along (C,H,W) dimensions. Instance normalization \cite{ulyanov2016instance} computes the mean and standard deviation along (H,W) dimensions. When the batch size is 1, batch normalization is equivalent to instance normalization.
Group normalization \cite{wu2018group} is a intermediate state between layer normalization and instance normalization. It uses a group of channels to compute the mean and standard deviation,
while layer normalization uses all channels, and instance normalization uses one channel.

Weight normalization \cite{salimans2016weight} normalizes the filter weights instead of the activations by re-parameterizing the incoming weight vector.  
Cosine normalization \cite{luo2017cosine} normalizes both the filter weights and the activations by using cosine similarity or Pearson correlation coefficient instead of dot product in neural networks.

Kalman normalization \cite{wang2018kalman} estimates the mean and standard deviation of a certain layer by considering the distributions of all its preceding layers, mimicking the merits of Kalman Filtering Process. It takes much overhead since it introduces a transition matrix and a covariance matrix to carry out Kalman transform, and needs sampling according to the distribution of previous layer before Kalman transform. 
On the other hand, it can be combined with other normalization, e.g. batch Kalman normalization, group Kalman normalization. Extended batch normalization could also be equipped with Kalman normalization.

Some researches try to use other statistics instead of mean and standard deviation in normalization. Instead of the standard $L^{2}$ batch normalization, \cite{hoffer2018norm} uses normalization in $L^{1}$ and $L^{\infty}$ spaces, and shows that can improve numerical stability in low-precision implementations as well as provide computational and memory benefits. Generalized batch normalization \cite{yuan2019generalized} investigates a variety of alternative deviation measures for scaling and alternative mean measures for centering.

Batch-instance normalization \cite{nam2018batch} uses a learnable gate parameter to combine batch and instance normalization together, and switchable normalization \cite{luo2018differentiable}  uses learnable parameters to combine batch, instance and layer normalization.
Virtual batch normalization \cite{salimans2016improved} and spectral normalization \cite{miyato2018spectral} focus on the normalization in generative adversarial networks. 
Self-Normalizing \cite{klambauer2017self} focuses on standard feed-forward neural networks (fully-connected networks).
Recurrent batch normalization \cite{cooijmans2016recurrent} modifies batch normalization to use in recurrent networks. 
EvalNorm \cite{singh2019evalnorm} estimates corrected normalization statistics to use for batch normalization during evaluation. \cite{ren2016normalizing} provides a unifying view of the different normalization approaches. 
\cite{santurkar2018does}, \cite{luo2018towards} and \cite{bjorck2018understanding} try to explain how batch normalization works.

\section{Extended Batch Normalization}
We first describe some notations which will be used next. In the case of NCHW format feature, let $U$ denote the universal set of the features in the same layer, and $i=(i_N,i_C,i_H,i_W)$ refer to a 4D tensor indexing the features, N is the batch dimension, C is the channel dimension, H and W are the spatial height and width dimensions. A family of normalization can be formalized as:

\begin{equation} \label{norm_eq}
\widehat{x}_{i}=\frac{1}{\sigma_{S_{i}^{'}}}(x_{i}-\mu_{S_{i}})
\end{equation}
\begin{equation} \label{affine_eq}
y_{i}=\gamma \widehat{x}_{i} + \beta
\end{equation}

Where x is the feature computed by a layer, and i is an index. 
$\mu$ is the mean and $\sigma$ is the standard deviation (std). $S_{i}$ is the set of pixels from which the mean is computed, and $ S_{i}^{'}$ is the set of pixels from which the standard deviation is computed. $\gamma$ and $\beta$ are learned parameters of affine transform.

In extended batch normalization, the set $S_{i}$ and $S_{i}^{'}$ are defined as:

\begin{equation} \label{ebn_mean_eq}
S_{i}=\{k|k_{C}=i_{C}\} 
\end{equation}
\begin{equation} \label{ebn_std_eq}
S_{i}^{'}=\{k|k \in U\} 
\end{equation}
Where $i_{C}$ (and $k_{C}$) refer to the sub-index of i (and k) along the C dimension, $U$ is the universal set of the features.

The key difference between extended batch normalization and other normalization methods is that extended batch normalization computes the mean and the standard deviation from different set of pixels. As the same as batch normalization, extended batch normalization computes the mean along the (N, H, W) dimensions. However, it computes the standard deviation along the (N,C,H,W) dimensions, thus enlarges the pixel set from which the standard deviation is computed.

\begin{figure}[tb]
\centering
\subfigure[BN]{
\label{FIGbnset}
\includegraphics[width=0.2\columnwidth]{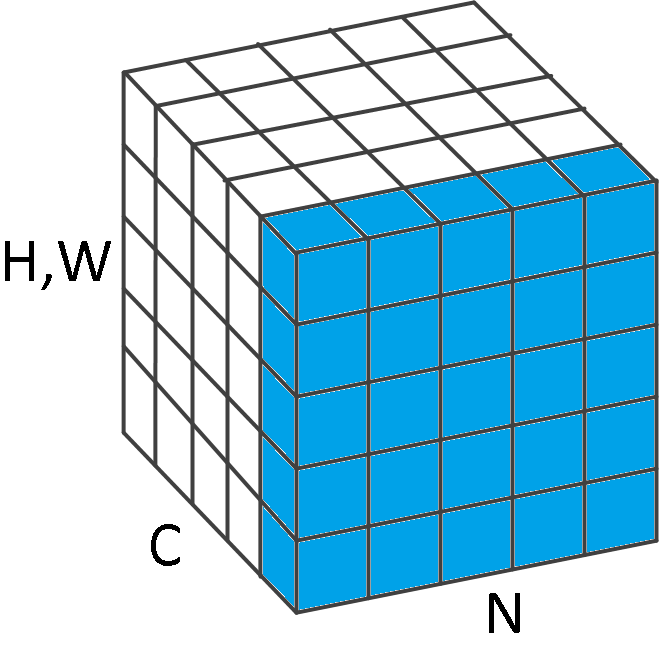}}
\centering
\subfigure[IN]{
\label{FIGlnset}
\includegraphics[width=0.2\columnwidth]{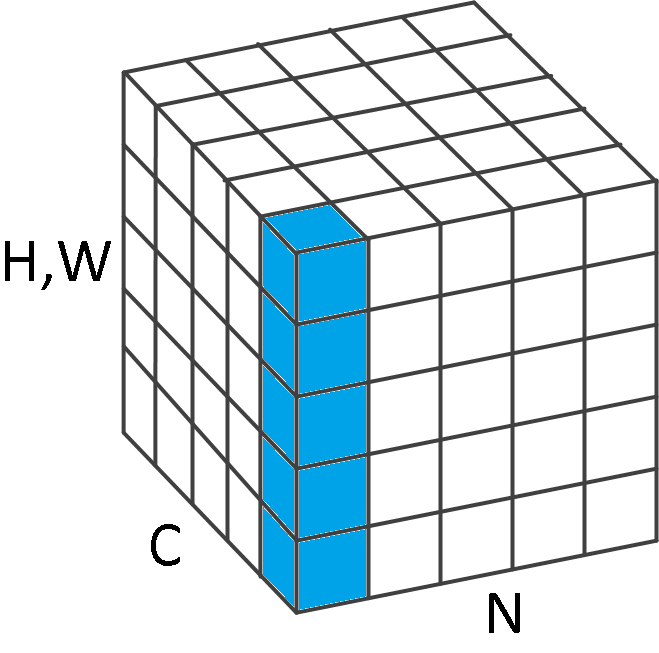}}
\centering
\subfigure[LN]{
\label{FIGlnset}
\includegraphics[width=0.2\columnwidth]{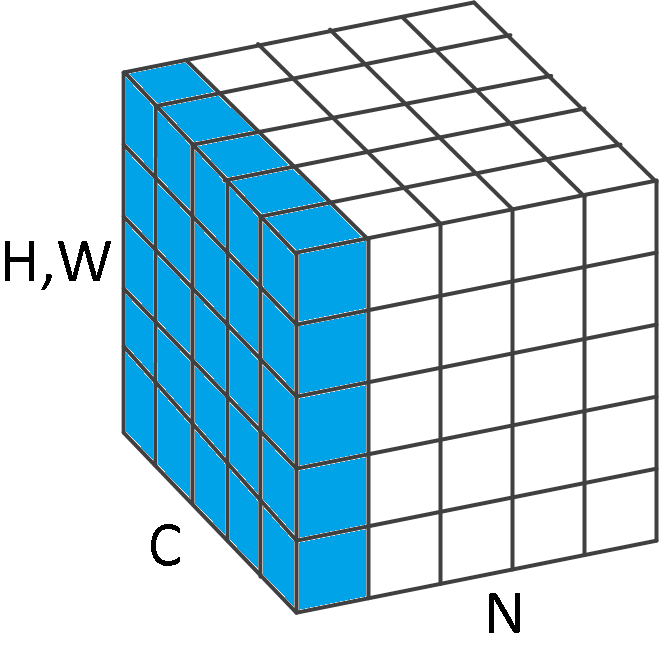}}
\centering
\subfigure[GN]{
\label{FIGgnset}
\includegraphics[width=0.2\columnwidth]{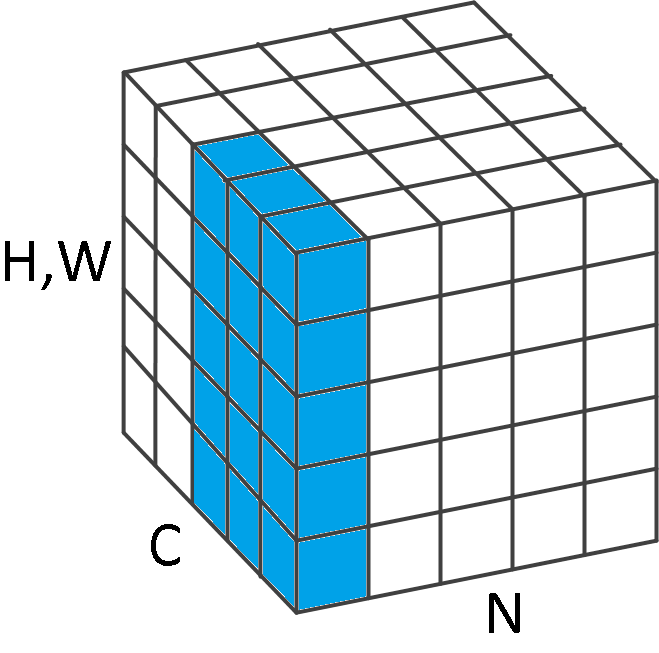}}
\centering
\subfigure[EBN (mean)]{
\label{FIGebnset1}
\includegraphics[width=0.2\columnwidth]{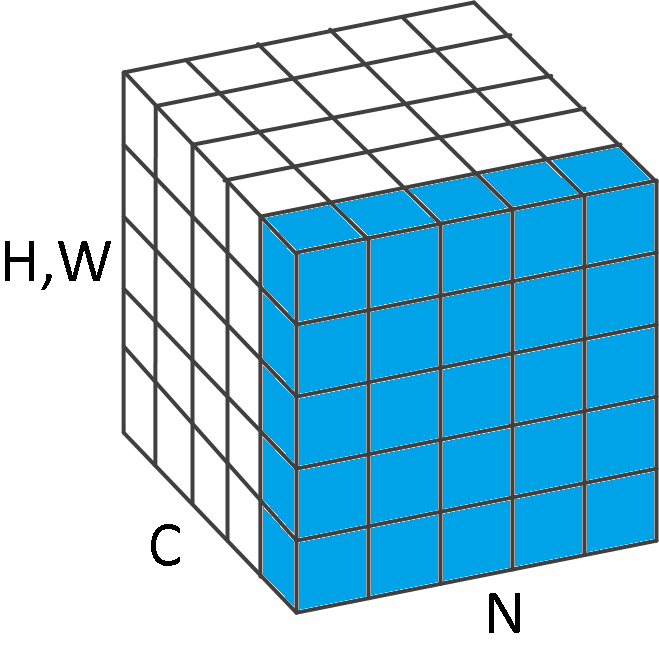}}
\hspace{.3in}
\centering
\subfigure[EBN (std)]{
\label{FIGebnset2}
\includegraphics[width=0.2\columnwidth]{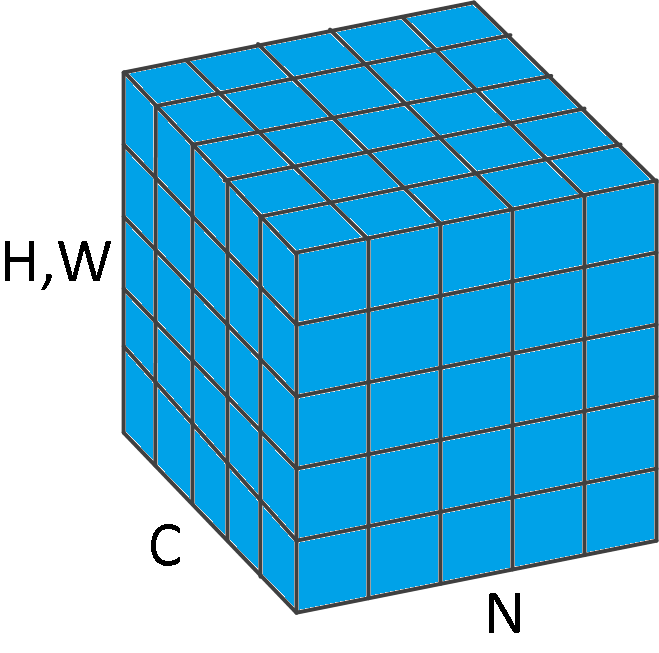}}
\caption{Different normalization methods. N refers to the batch dimension, C refers to the channel dimension, and (H, W) refers to the spatial dimension. The pixels in blue are the pixel set from which the  mean and the standard deviation (std) are computed. (a) Batch norm.  (b) Instance norm .(c) Layer norm. (d) Group norm. (e) Extended batch norm (mean). (f) Extended batch norm (std). In extended batch normalization, the mean and the standard deviation are computed from different pixel sets}
\label{diffnorm}
\end{figure}

The different normalization methods are shown in Fig \ref{diffnorm}. Batch normalization computes both the mean and the standard deviation along the (N, H, W) dimensions (Equation \ref{bn_eq}). Layer normalization computes both the mean and the standard deviation along the (C, H, W) dimensions (Equation \ref{ln_eq}).  Instance normalization computes both the mean and the standard deviation along the (H, W) dimensions (Equation \ref{in_eq}). Group normalization computes both the mean and the standard deviation in (H, W) dimensions and along a group of $C/G$ channels where G is the number of channel groups (Equation \ref{gn_eq}). 

\begin{equation} \label{bn_eq}
S_{i}=S_{i}^{'}=\{k|k_{C}=i_{C}\} 
\end{equation}
\begin{equation} \label{ln_eq}
S_{i}=S_{i}^{'}=\{k|k_{N}=i_{N}\} 
\end{equation}
\begin{equation} \label{in_eq}
S_{i}=S_{i}^{'}=\{k|k_{C}=i_{C},k_{N}=i_{N}\} 
\end{equation}
\begin{equation} \label{gn_eq}
S_{i}=S_{i}^{'}=\{k|\frac{k_{C}}{C/G}=\frac{i_{C}}{C/G},k_{N}=i_{N}\} 
\end{equation}

In the case of NC format input, e.g. input in fully-connected networks, extended batch normalization computes the mean along the (N) dimension, and computes the standard deviation along the (N, C) dimensions.

\subsection{Small Batch Size}
When training with batch normalization, examples are seen in conjunction with each other in the minibatch. The model looks at multiple training examples in combination, rather than in isolation. The coordination between examples helps the learning process. Moreover, the random selection of examples in the minibatch brings the sampling noises, and that provides similar regularization benefits as Dropout \cite{srivastava2014dropout}. However, when the batch size is small, the noises are too much and make inaccurate batch statistics estimation.

Let $m_{N}$ refer to the size of N dimension (batch size), $m_{C}$ refer to the size of C dimension, $m_{H}$ refer to the size of H dimension, $m_{W}$ refer to the size of W dimension, $m$ refer to the size of $S_{i}$. Then, in batch normalization, 
\begin{equation} \label{bn_m_eq}
m=m_{N}*m_{H}*m_{W}
\end{equation}
When $m_{N}$ is small, the $m$ is small. That is to say, the pixel set from which the mean and the standard deviation are computed is small. That makes the estimation of mean and standard deviation inaccurate. In layer normalization,
\begin{equation} \label{ln_m_eq}
m=m_{C}*m_{H}*m_{W}
\end{equation}
The $m$ in layer normalization is independent on $m_{N}$, thus has no problem caused by small batch size. However, normalizing along the channel dimension increases the correlation between the channels since they share the same mean and standard deviation. Correlation between features is harmful for training neural networks \cite{lecun2012efficient}. Group normalization alleviates this problem by normalizing along a group of channels. The size $m$ in group normalization is:
\begin{equation} \label{ln_m_eq}
m=\frac{m_{C}}{G}*m_{H}*m_{W}
\end{equation}
In the same group, the correlation between the channels still exists. Moreover, there is a extra hyper parameter $G$ to be tuned. When $G$ is large, the $m$ is small. Then the estimation of mean and standard deviation may be inaccurate. As $G$ equals to $m_{C}$, group normalization becomes instance normalization. When $G$ is small, the number of channels in the same group increase, but this aggravate the problem of correlation. When $G$ equals to 1, group normalization becomes layer normalization. Lastly, for low resolution input, the $m_{H}*m_{W}$ is small, that also makes the estimation of the mean and the standard deviation  inaccurate. The extreme situation is $m_{H}*m_{W}=1$, the input becomes 1-dimension vector.

In extended batch normalization, the mean is computed along (N, H, W) dimensions as the same as batch normalization. Thus it maintains the merits of batch normalization. On the other hand, extended batch normalization enlarges the pixel set $S_{i}^{'}$ from which  the standard deviation is computed along (N, C, H, W) dimensions,  thus alleviates the problem caused by small batch size or low resolution.  The size of $S_{i}^{'}$ is: 
\begin{equation} \label{ln_m_eq}
m^{'}=m_{N}*m_{C}*m_{H}*m_{W}
\end{equation}

\subsection{Inference}
At inference time for extended batch normalization, the mean and the standard deviation are pre-computed
from the training set by the moving average as the same as batch normalization. 

\begin{equation} \label{infer_eq}
\begin{aligned}
y_i &=\gamma (\frac{1} {\sigma_{r}^{t}} (x_i- \mu_{r}^{t} ))+ \beta \\
     &=\frac{\gamma }{\sigma_{r}^{t}} x_i + (\beta - \frac{\gamma \mu_{r}^{t}}{\sigma_{r}^{t}})
\end{aligned}
\end{equation}

\begin{equation} \label{running_mean_eq}
\mu_{r}^{t} = (1-\rho)\mu_{r}^{t-1} + \rho \mu_{b}^{t}
\end{equation}

\begin{equation} \label{running_std_eq}
\sigma_{r}^{t} = (1-\rho)\sigma_{r}^{t-1} + \rho \sigma_{b}^{t}
\end{equation}

Where $\mu_{r}$ refers to the moving average of the mean,  $\sigma_{r}$  refers to the moving average of the standard deviation, $\mu_{b}$ refers to the mean of a batch, $\sigma_{b}$ refers to the standard deviation of a batch,  $t$ refers to the number of times of computation, and $\rho$ is a momentum constant less than one.  

There is no need to compute the mean and the standard deviation at inference time comparing to group normalization.  Moreover, since the mean and the standard deviation are pre-computed and fixed at inference time, the normalization can be fused into convolution operation. That is very helpful to speed up the inference especially on the mobile or embedded devices.

\section{Experiment}
In this section, we compare extended batch normalization (EBN) with batch normalization (BN) and group normalization (GN) on the datasets of MNIST, CIFAR-10/100, STL-10, and ImageNet, respectively. We use Pytorch 1.1.0 in our experiments $\footnote{The codes for MNIST and ImageNet experiments are based on Pytorch examples (https://github.com/pytorch/examples), and CIFAR is based on pytorch-cifar (https://github.com/kuangliu/pytorch-cifar). }$. BN and GN have already been implemented by Pytorch. For BN, we use the default settings in Pytorch. For EBN, we use the same settings as BN. For GN, we set the group G to 32 which is used in the original paper \cite{wu2018group}.

\subsection{MNIST}
The MNIST \cite{lecun1998gradient} data set consists of 28x28 pixel  handwritten digit black and white images. The task is to classify the images into 10 digit classes. There are  60, 000 training  images and 10, 000 test images in the MNIST data set. Each channel of the input is normalized into 0 mean and 1 std globally. 

Our purpose is to compare different normalization methods, rather than renew the record. We used a very simple network, with a 28x28 binary image as input, and 4 fully-connected hidden layers with 128 hidden units each. Normalization followed by ReLU activation is used each layer. 
We evaluate the batch sizes of 128 and 4 with different normalization. For the batch size of 128, the learning rate is set to 0.1. And for the batch size of 4, the learning rate is set to $0.1*4/128$, following the linear scaling rule \cite{goyal2017accurate} to adapt to batch size changes. We train the network for 50 epochs using SGD with 0.5 momentum. We use 1 GPU to train all models.

\begin{figure}[!htb]
\centering
\subfigure[Batch size is 128]{
\label{fig_mnist_a}
\includegraphics[width=0.4\columnwidth]{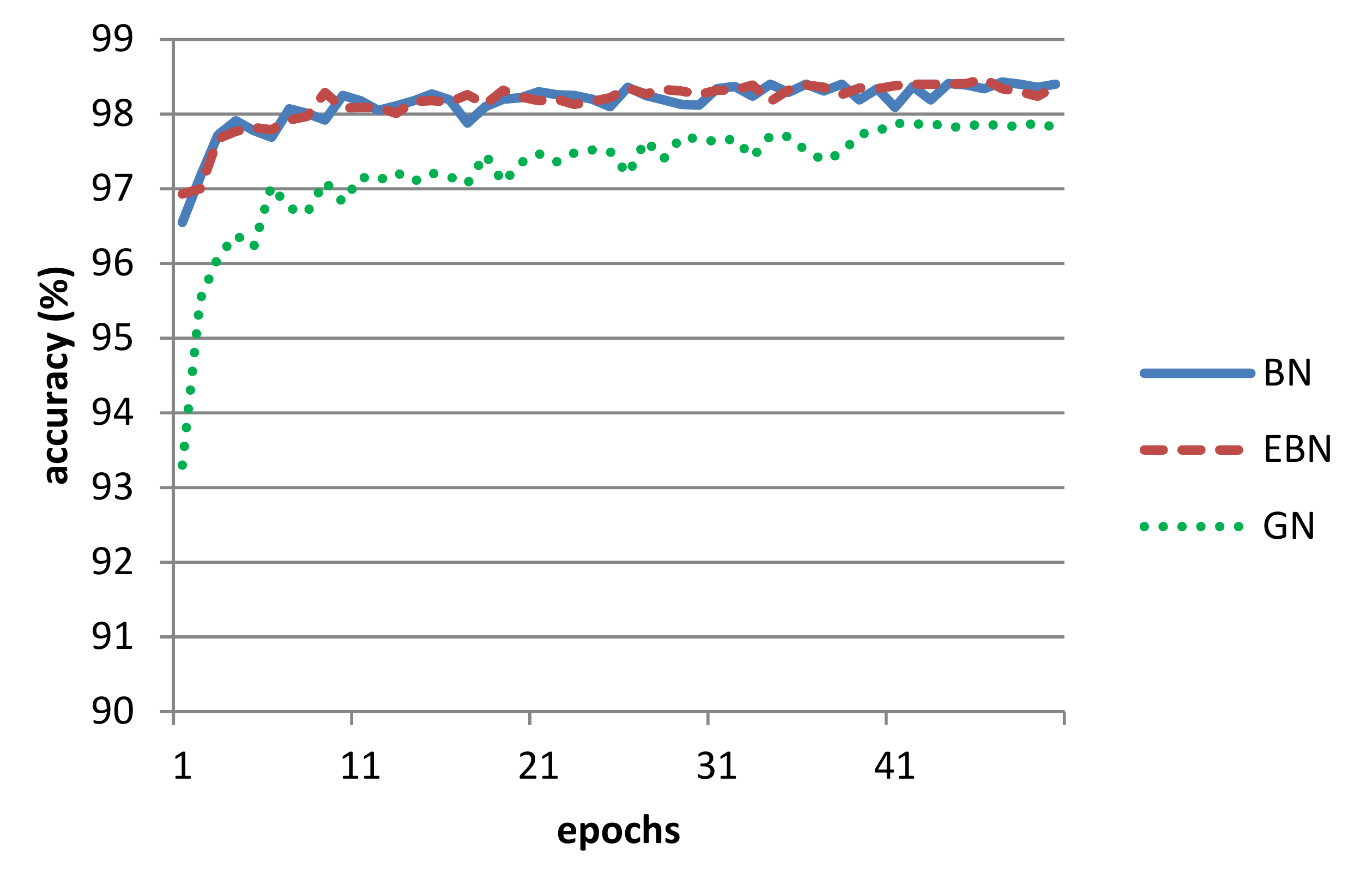}}
\centering
\subfigure[Batch size is 4]{
\label{fig_mnist_b}
\includegraphics[width=0.4\columnwidth]{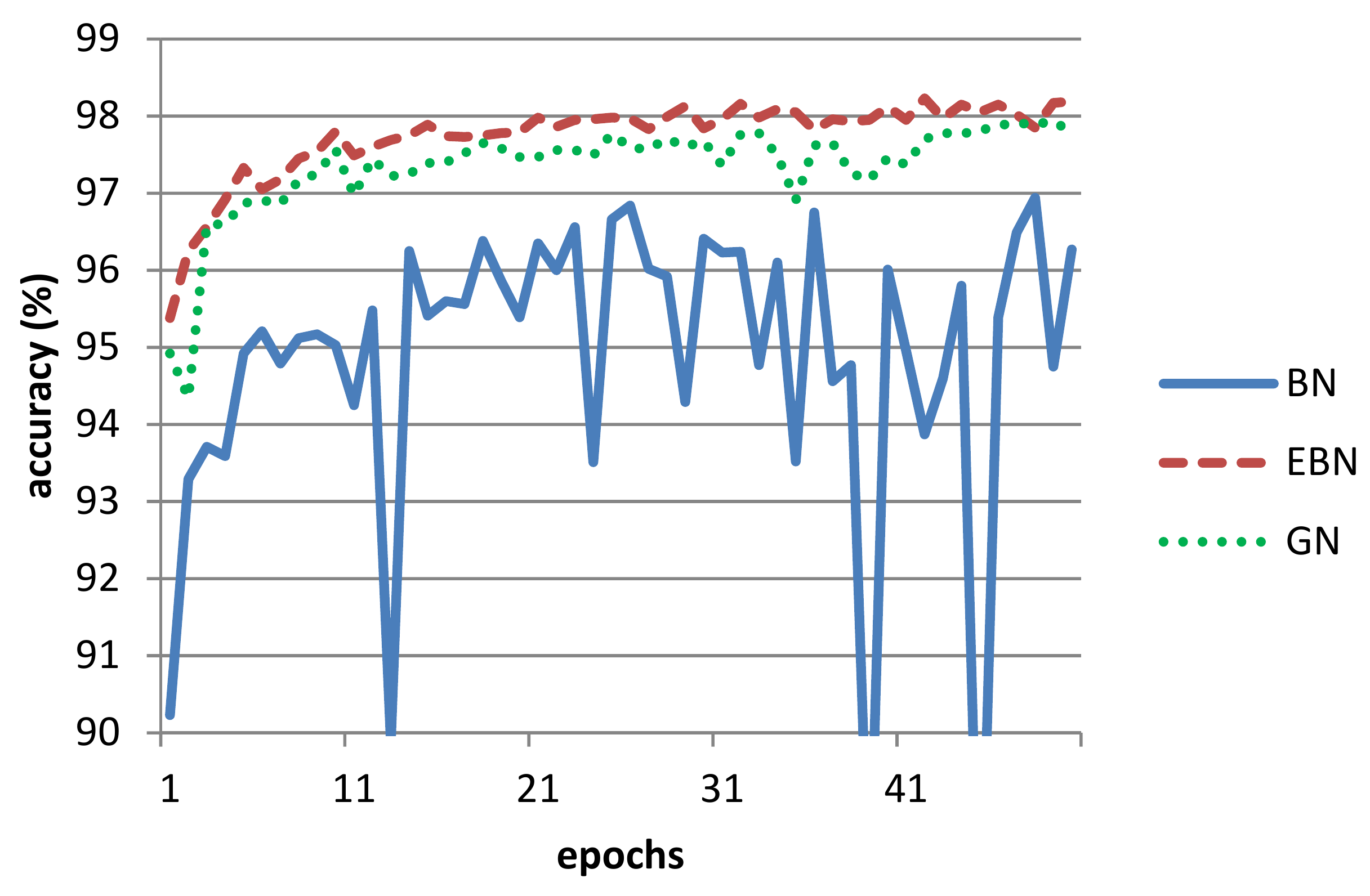}}
\caption{The test accuracy of the fully-connected network on MNIST vs. the number of training epoch, with different normalization methods}
\label{fig_mnist}
\end{figure}

The results of MNIST are shown in Fig \ref{fig_mnist}  and Table \ref{tab_mnist}. For the batch size of 128, shown in Fig \ref{fig_mnist_a}, we can see BN and EBN achieve similar performances which are better than GN. For the batch size of 4, shown in Fig \ref{fig_mnist_b}, the test accuracy of BN has large fluctuation. EBN achieves the most stable and highest accuracy. We show the accuracy numbers in Table \ref{tab_mnist}. To reduce random variations, we report the average accuracy of the final 5 epochs (This is adopted in our following experiments). With the batch size of 128, EBN achieves 98.37\% test accuracy, 0.52\% higher than GN, which achieves 97.85\%. With the batch size of 4, the test accuracies of all normalization methods have decreased. 
Compared to the batch size of 128, BN, EBN and GN reduces 3.76\%, 0.3\% and 0.07\% respectively. EBN is still better than GN by 0.29\%.  

\begin{table}[!htb]
\caption{The test accuracy (\%) of the fully-connected network on MNIST}
\label{tab_mnist}
\centering
\begin{tabular}{l|lll}
\hline
  & BN & EBN & GN \\
\hline
batch size=128     & 98.33  & 98.37 & 97.85     \\
batch size=4       & 94.57  & 98.07 & 97.78     \\
\hline
\end{tabular}
\end{table}

\subsubsection{Summary} In the case of fully-connected network,  $n_{H}*n_{W}=1$. The size of pixel set, from which the statistics are computed for batch normalization,  depends only on the batch size. Thus the performance of BN is very poor when using small batch size. EBN maintains the advantage of BN in case of the large batch size. Meanwhile, it alleviates the problem caused by small batch size. 

\subsection{CIFAR}

CIFAR-10 \cite{krizhevsky2009learning} is a data set of natural 32x32 RGB images in 10 classes with 50, 000 images for training and 10, 000 for testing. CIFAR-100 is similar with CIFAR-10 but with 100 classes. To augment data, the training images are padded with 0 to 36x36 and then randomly cropped to 32x32 pixels. Then randomly horizontal flipping is made. Each channel of the input is normalized into 0 mean and 1 std globally. Weight decay of 0.0005, and SGD with 0.9 momentum are used. 

On CIFAR-10 and CIFAR-100, We evaluate both ResNet18 \cite{he2016deep} and VGG16 \cite{simonyan2014very} with BN, EBN and GN, respectively. For the batch size of 128, the initial learning rate is set to 0.1. And for the batch size of 4, the initial learning rate is set to $0.1*4/128$, following the linear scaling rule. We train the network for 200 epochs, and decrease the learning rate by 10x at 100 and 150 epochs.. The training of VGG16 with GN is failed when using the initial learning rate of 0.1, thus we decrease the initial learning rate to 0.01 (we also try the initial learning rate of 0.05, but the training is still failed) when training the VGG16 with GN. We use 1 GPU to train all models.

\subsubsection{Results of ResNet18}
\begin{figure}[!htb]
\centering
\subfigure[Batch size is 128]{
\label{fig_cifar10_a}
\includegraphics[width=0.4\columnwidth]{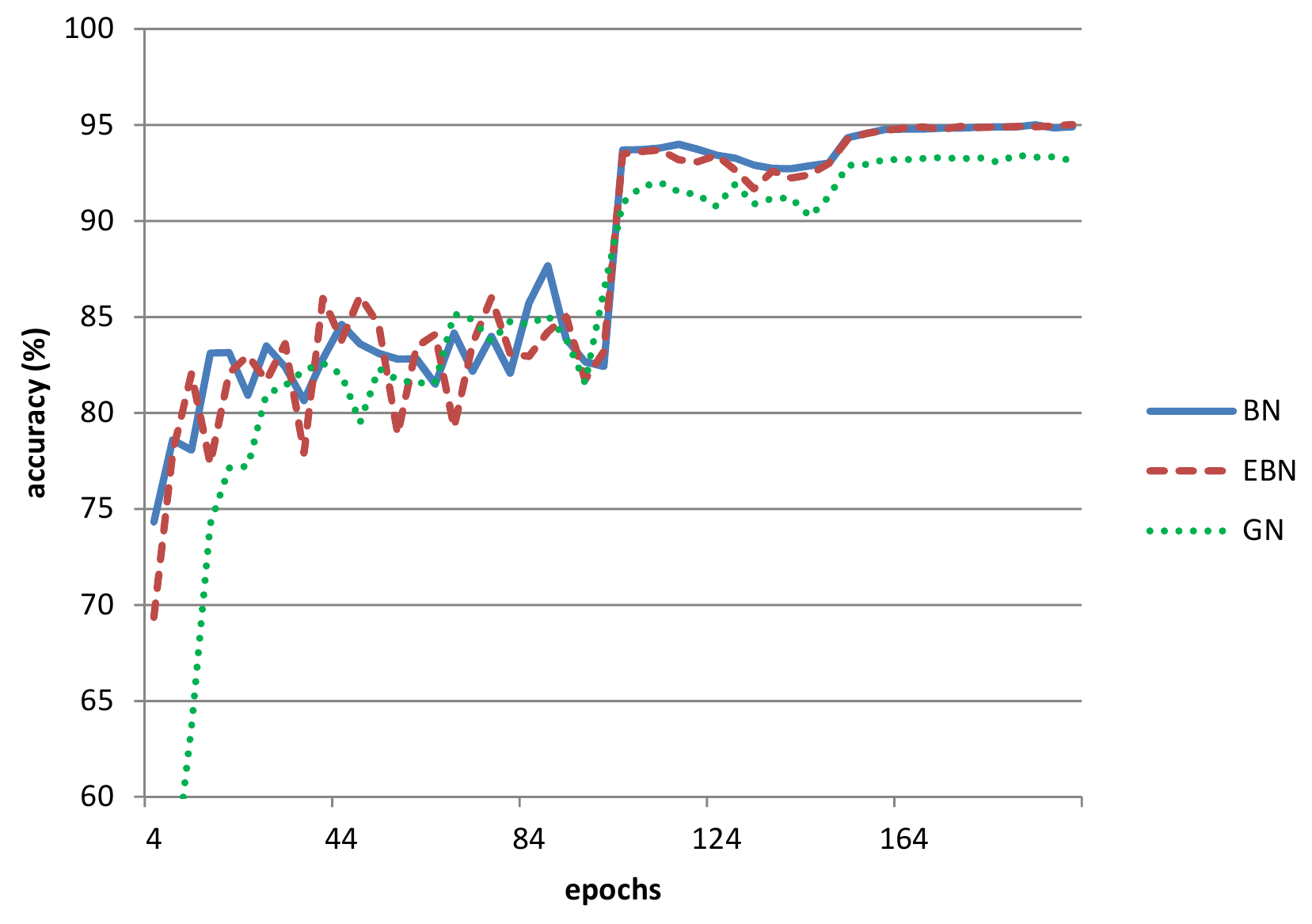}}
\centering
\subfigure[Batch size is 4]{
\label{fig_cifar10_b}
\includegraphics[width=0.4\columnwidth]{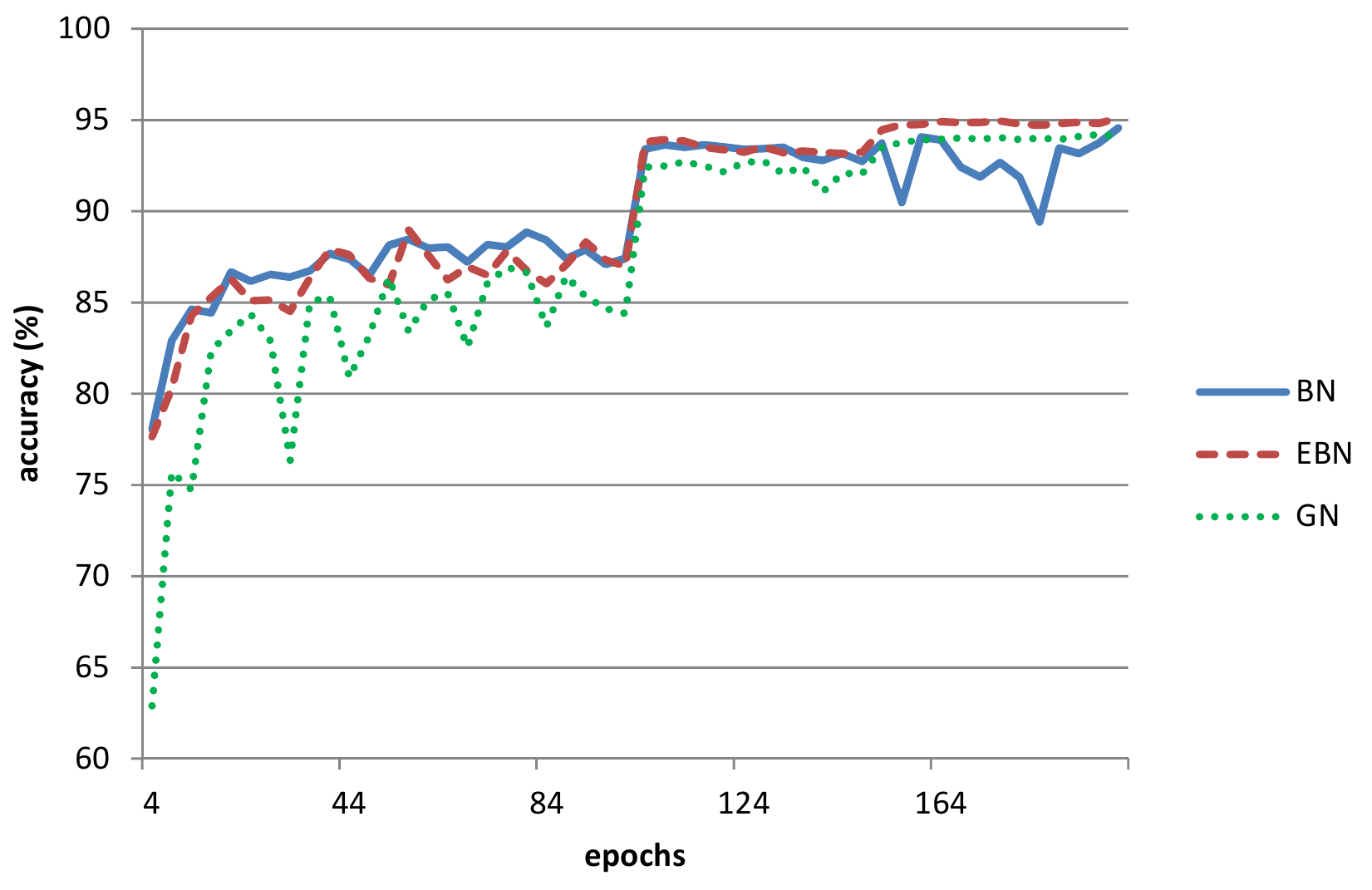}}
\caption{The test accuracy of ResNet18 on CIFAR-10 vs. the number of training epoch, with different normalization methods}
\label{fig_cifar10}
\end{figure}

The results of ResNet18 on CIFAR-10 are shown in Fig \ref{fig_cifar10}  and Table \ref{tab_cifar10}. From Fig \ref{fig_cifar10_a}, we can see BN and EBN achieve close performances with the large batch size of 128. EBN has 94.96\%  test accuracy, slightly better than BN, and better than GN by 1.62\%. As shown is Fig \ref{fig_cifar10_b}, with the small size of 4, BN  also has large fluctuation of test accuracy, while GN achieves better performance than with the batch size of 128.  Stably, EBN achieves the best test accuracy 94.92\% in the case of the small batch size.

\begin{table}[!htb]
\caption{The test accuracy (\%) of ResNet18 on CIFAR-10}
\label{tab_cifar10}
\centering
\begin{tabular}{l|lll}
\hline
  & BN & EBN & GN \\
\hline
batch size=128     & 94.92  & 94.96 & 93.34     \\
batch size=4       & 91.94  & 94.92 & 94.18     \\
\hline
\end{tabular}
\end{table}

\begin{figure}[!htb]
\centering
\subfigure[Batch size is 128]{
\label{fig_cifar100_a}
\includegraphics[width=0.4\columnwidth]{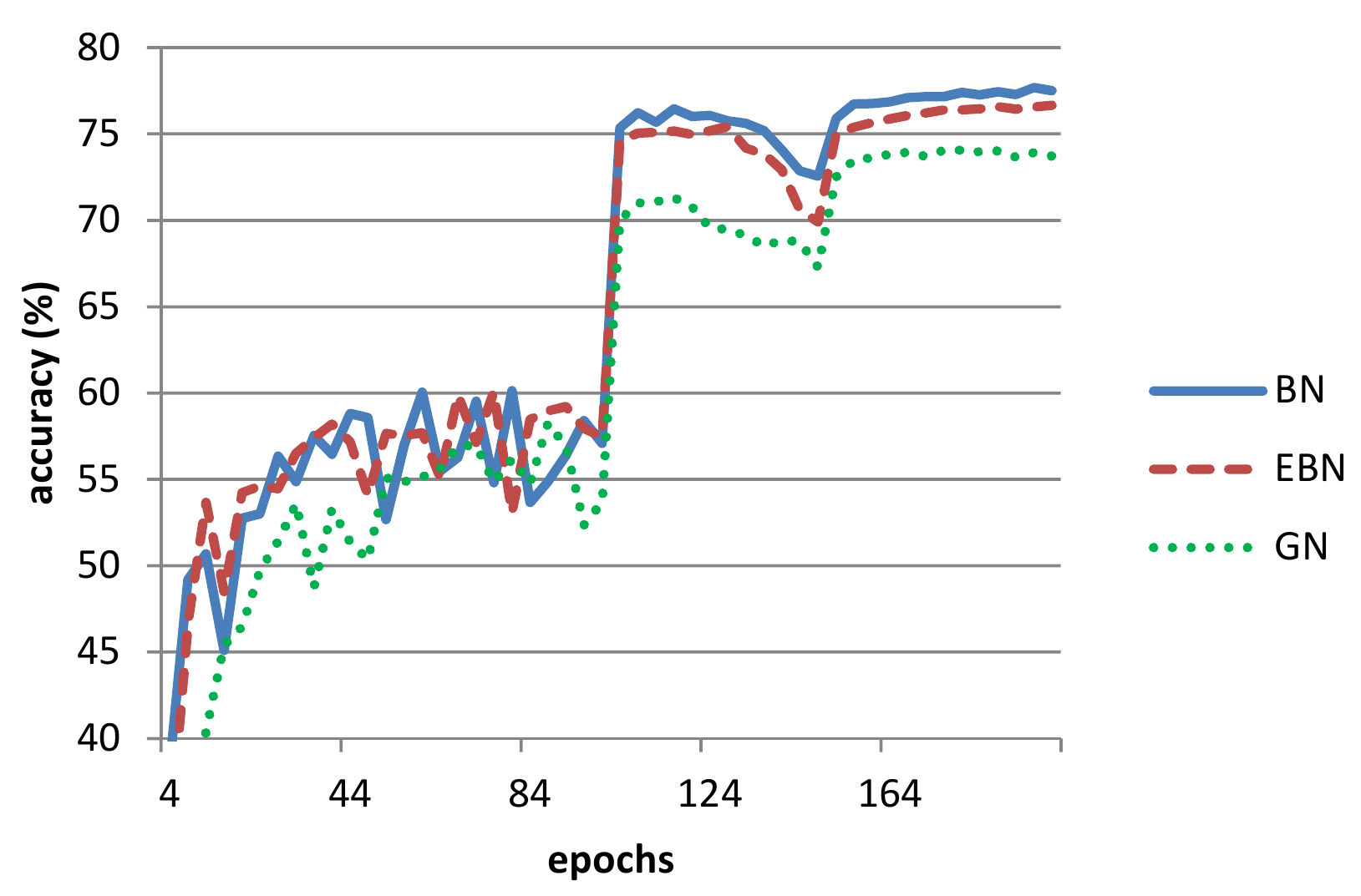}}
\centering
\subfigure[Batch size is 4]{
\label{fig_cifar100_b}
\includegraphics[width=0.4\columnwidth]{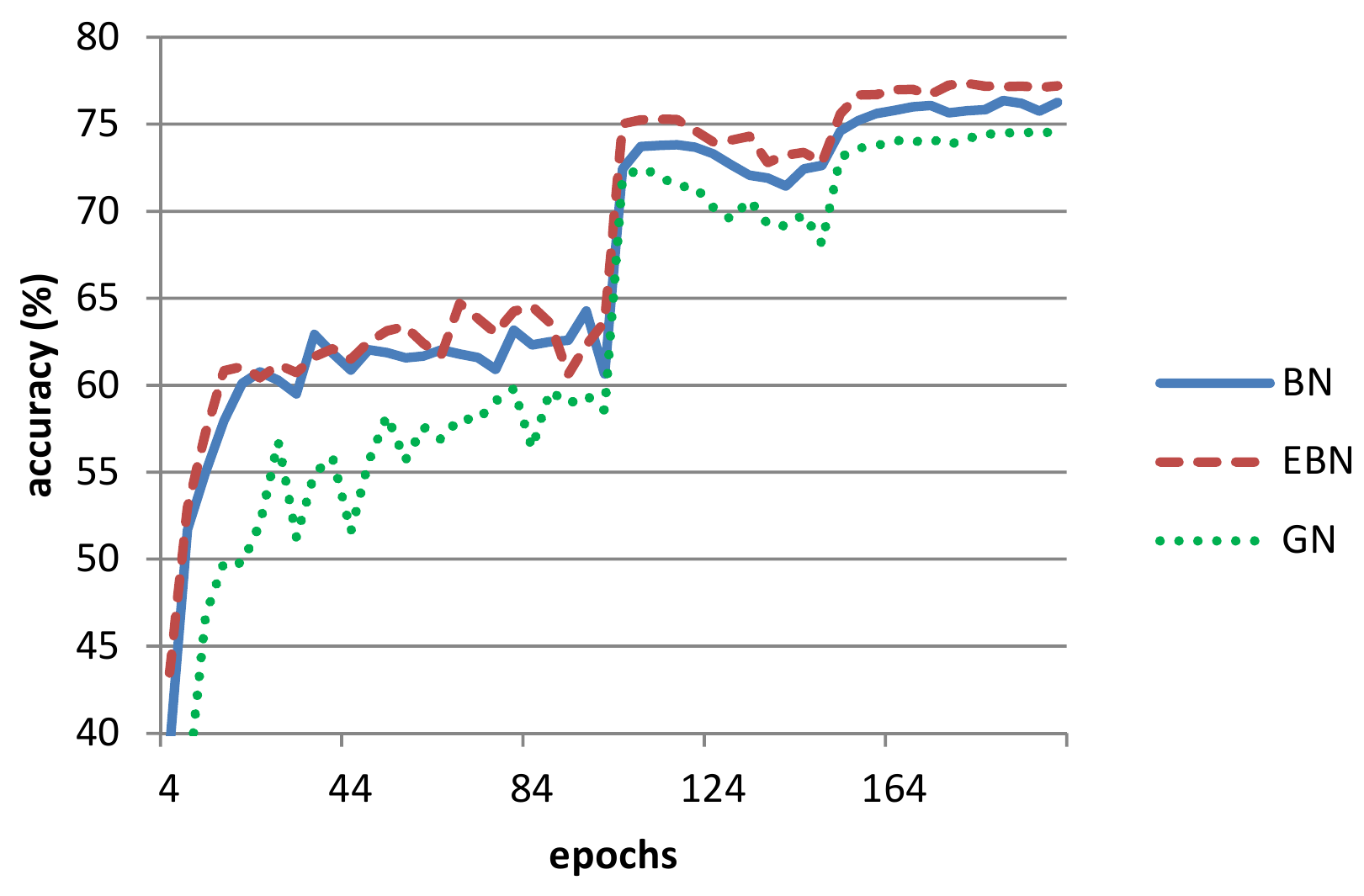}}
\caption{The test accuracy of ResNet18 on CIFAR-100 vs. the number of training epoch, with different normalization methods}
\label{fig_cifar100}
\end{figure}

The results of ResNet18 on CIFAR-100 are shown in Fig \ref{fig_cifar100}  and Table \ref{tab_cifar100}.
With the batch size of 128, BN achieves the best test accuracy of 77.50\%. EBN has 76.53\%, worse than BN by 0.97\%, but better than GN by 2.68\% which only has 73.85\%. When the batch size decreases to 4, BN' performance decreases, while EBN' and GN' performance increase. EBN achieves the best test accuracy 77.17\%, better than BN by 1.32\%, and GN by 2.71\%.

\begin{table}[!htb]
\caption{The test accuracy (\%) of ResNet18 on CIFAR-100}
\label{tab_cifar100}
\centering
\begin{tabular}{l|lll}
\hline
  & BN & EBN & GN \\
\hline
batch size=128     & 77.50  & 	76.53 & 73.85     \\
batch size=4       & 75.85  & 77.17 & 74.46     \\
\hline
\end{tabular}
\end{table}

\subsubsection{Results of VGG16}

\begin{figure}[!htb]
\centering
\subfigure[Batch size is 128]{
\label{fig_cifar10_vgg16_a}
\includegraphics[width=0.4\columnwidth]{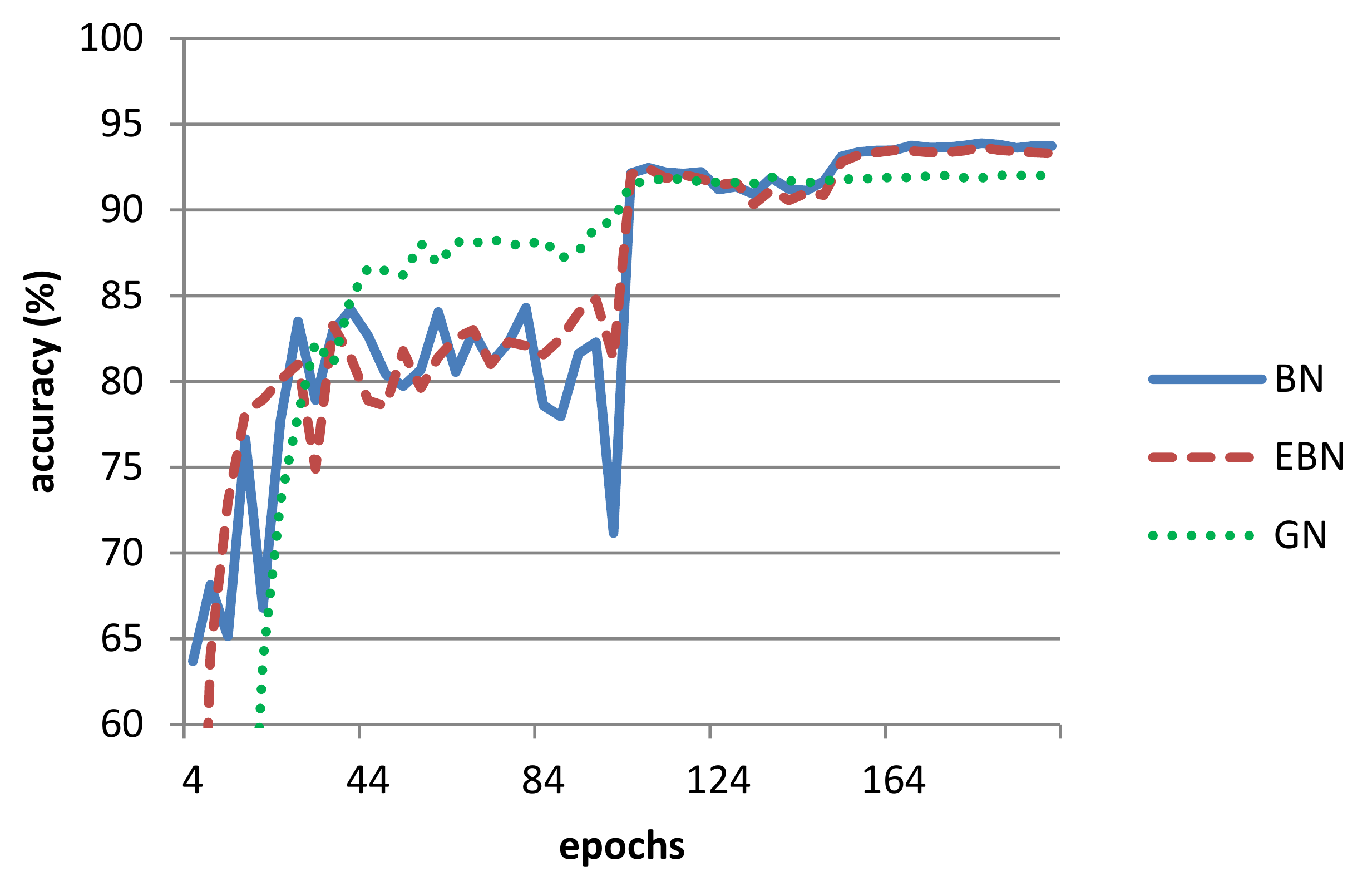}}
\centering
\subfigure[Batch size is 4]{
\label{fig_cifar10_vgg16_b}
\includegraphics[width=0.4\columnwidth]{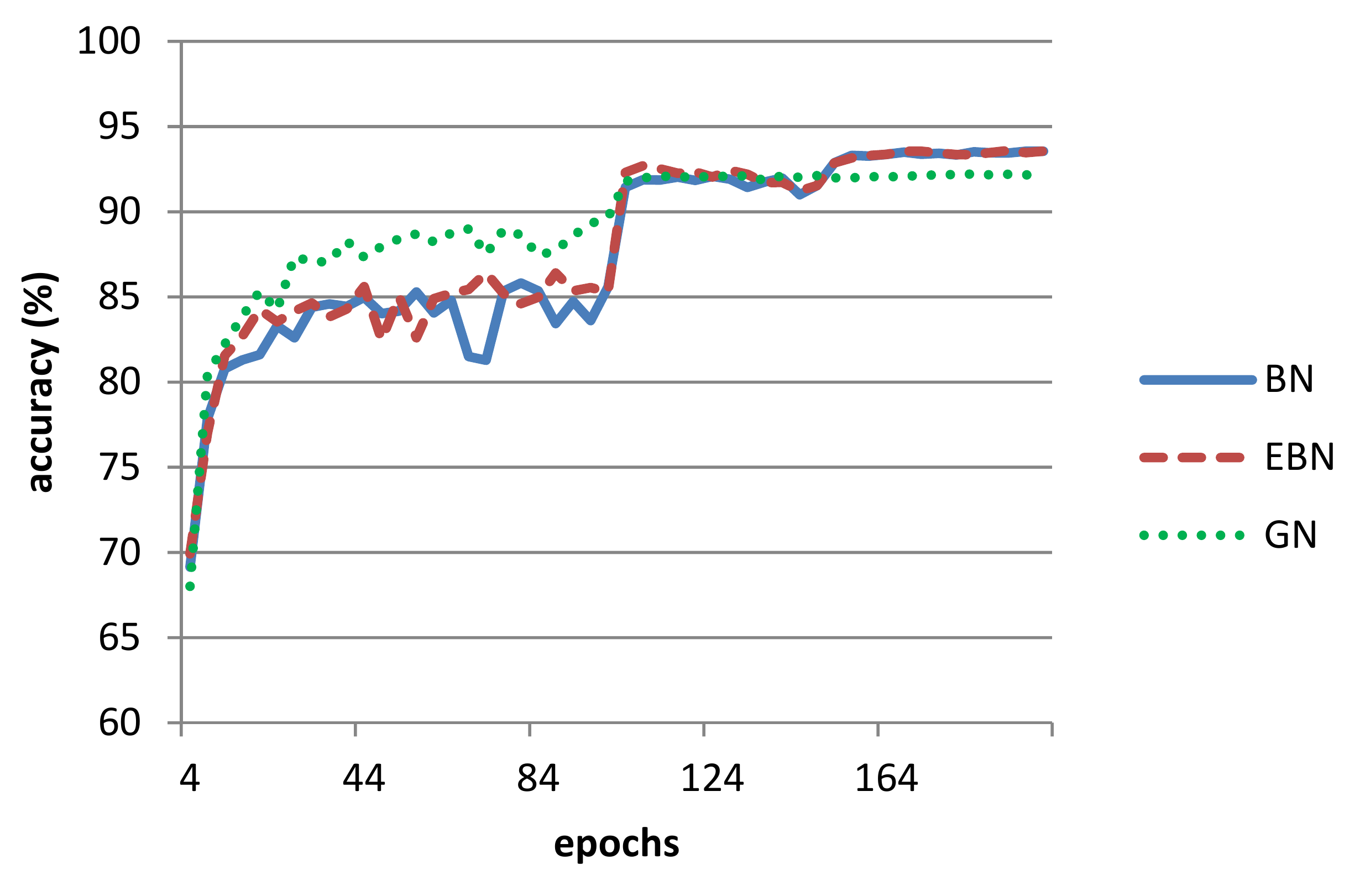}}
\caption{The test accuracy of VGG16 on CIFAR-10 vs. the number of training epoch, with different normalization methods}
\label{fig_cifar10_vgg16}
\end{figure}

The results of VGG16 on CIFAR-10 are shown in Fig \ref{fig_cifar10_vgg16}  and Table \ref{tab_cifar10_vgg16}.With the large batch size of 128, BN achieves best accuracy 93.71\%. EBN have 93.41\% test accuracy, worse than BN by 0.,3\%, but better than GN by 1.4\%.  
With the small size of 4, BN' accuracy decreases to 93.53\%, while EBN' increases to 93.55\%. They are better than the GN' accuracy 92.12\%.

\begin{table}[!htb]
\caption{The test accuracy (\%) of VGG16 on CIFAR-10}
\label{tab_cifar10_vgg16}
\centering
\begin{tabular}{l|lll}
\hline
  & BN & EBN & GN \\
\hline
batch size=128     & 93.71  & 93.41 & 92.0     \\
batch size=4       & 93.53  & 93.55 & 92.12     \\
\hline
\end{tabular}
\end{table}

\begin{figure}[!htb]
\centering
\subfigure[Batch size is 128]{
\label{fig_cifar100_vgg16_a}
\includegraphics[width=0.4\columnwidth]{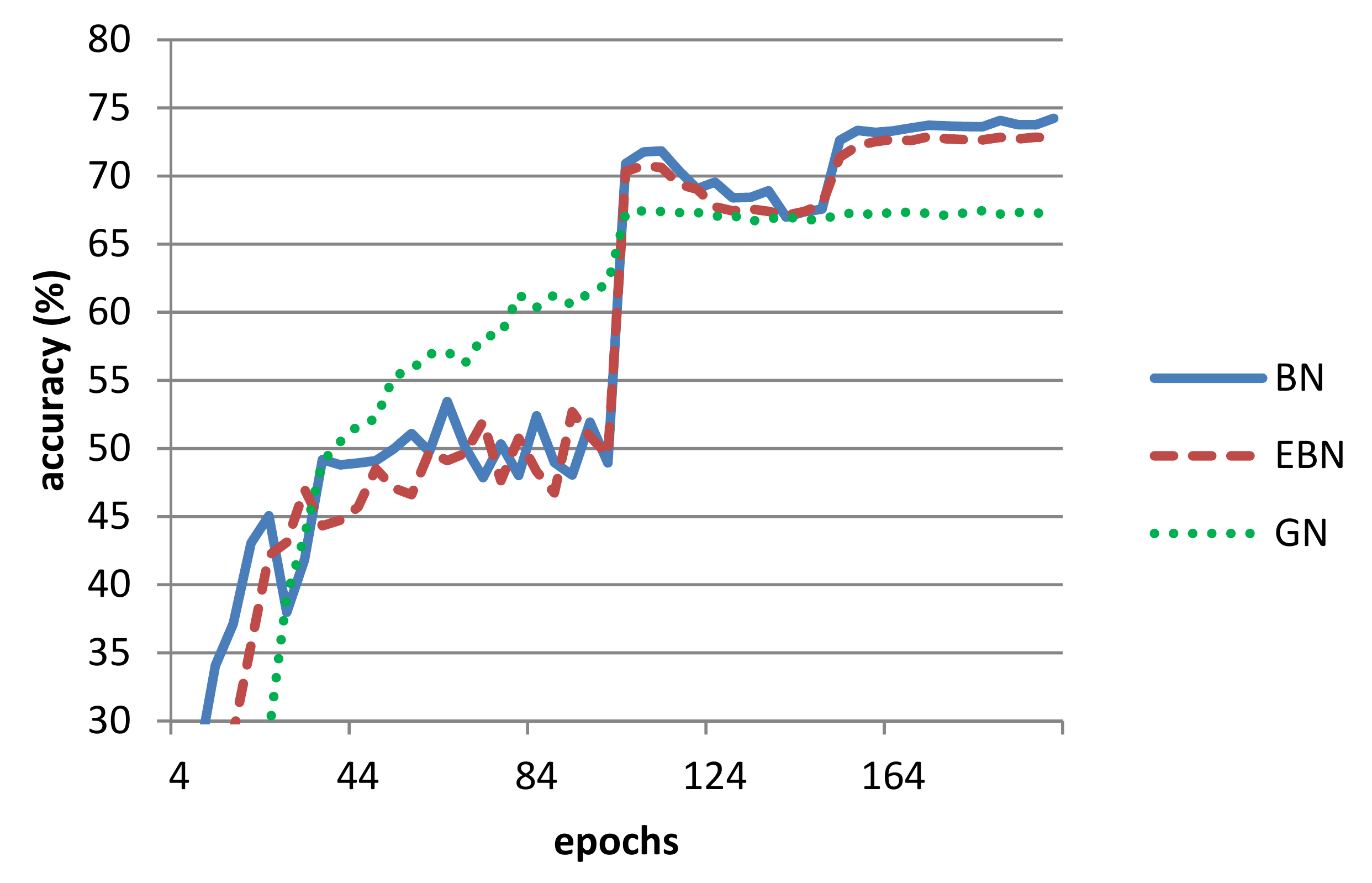}}
\centering
\subfigure[Batch size is 4]{
\label{fig_cifar100_vgg16_b}
\includegraphics[width=0.4\columnwidth]{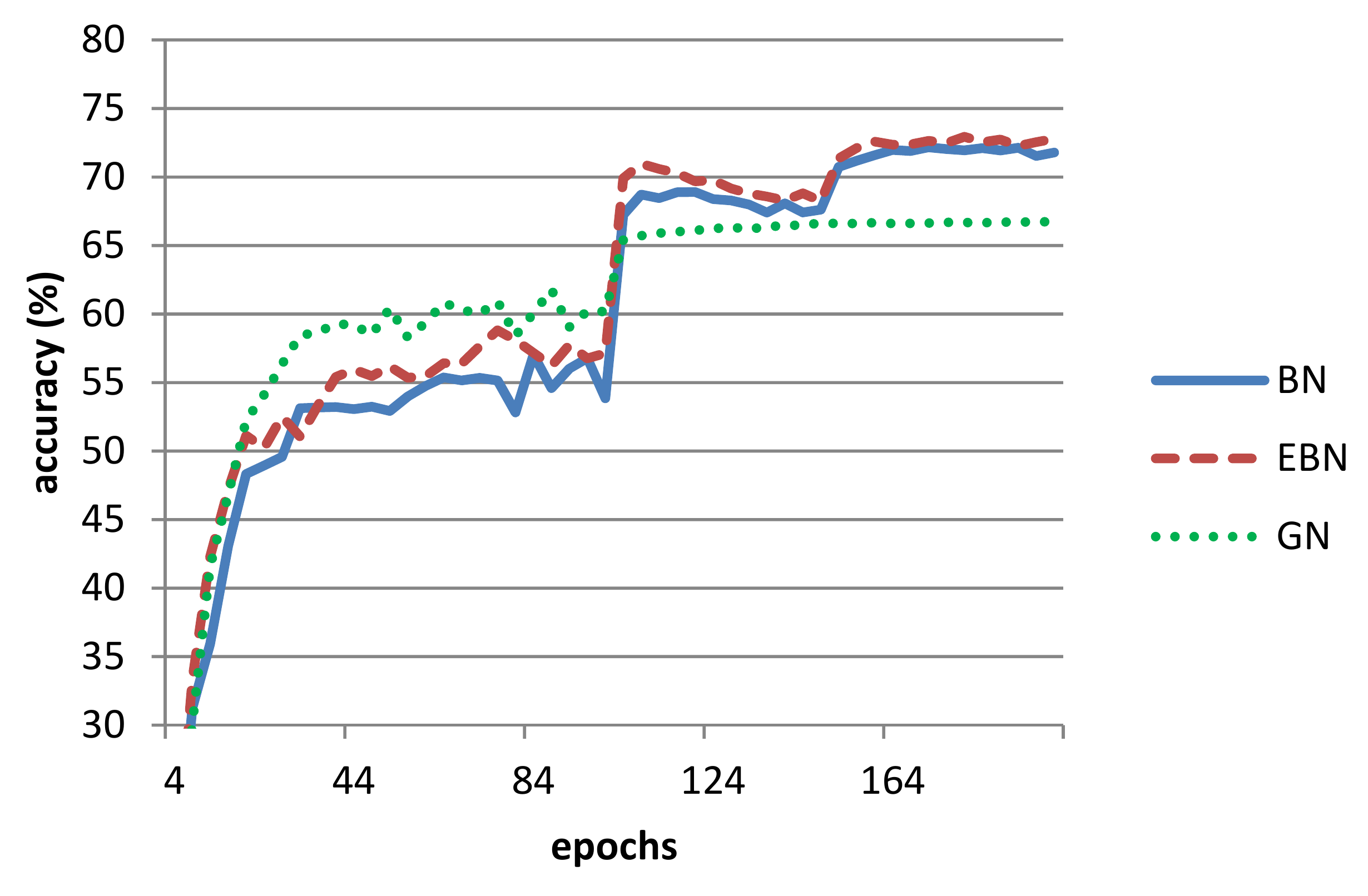}}
\caption{The test accuracy of VGG16 on CIFAR-100 vs. the number of training epoch, with different normalization methods}
\label{fig_cifar100_vgg16}
\end{figure}

The results of VGG16 on CIFAR-100 are shown in Fig \ref{fig_cifar100_vgg16}  and Table \ref{tab_cifar100_vgg16}.
With the batch size of 128, BN achieves the best test accuracy of 74.03 \%. EBN has 72.80\%, and GN  only has 67.29\%. With the batch size of 4, EBN achieves the best test accuracy 72.66\%, better than BN by 0.93\%, and than GN by 5.94\%.

\begin{table}[!htb]
\caption{The test accuracy (\%) of VGG16 on CIFAR-100.}
\label{tab_cifar100_vgg16}
\centering
\begin{tabular}{l|lll}
\hline
  & BN & EBN & GN \\
\hline
batch size=128     & 74.03  & 	72.80 & 67.29     \\
batch size=4       & 71.73  & 72.66 &   66.72     \\
\hline
\end{tabular}
\end{table}

\subsubsection{Summary}  With the large batch size, BN works well both in ResNet18 and VGG16. EBN performs slightly worse than BN, but better than GN with the large batch size. Finally, EBN performs better than both BN and GN when using the small batch size. 

\subsection{STL-10}

The STL-10 dataset is a data set of natural 96x96 RGB images in 10 classes. Each class has fewer labeled training examples, but a very large set of unlabeled examples. We use only the labeled examples (500 training images, 800 test images per class) in our experiments. 
To augment data, the training images are padded with 0 to 100x100 and then randomly cropped to 96x96 pixels. We evaluate ResNet18 with different normalization methods. The experiment settings are the same as CIFAR-10.

\begin{figure}[!htb]
\centering
\subfigure[Batch size is 128]{
\label{fig_stl_a}
\includegraphics[width=0.4\columnwidth]{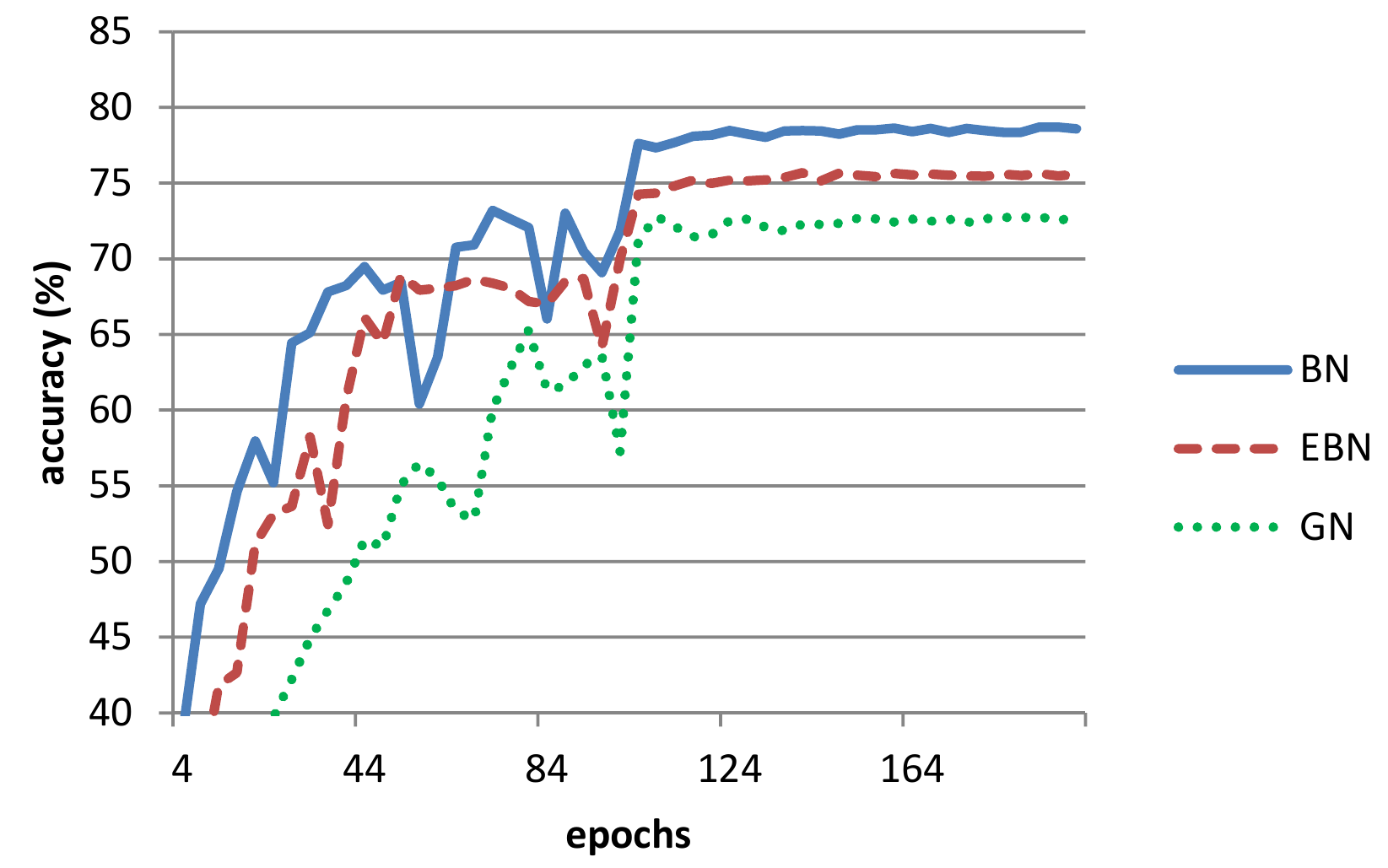}}
\centering
\subfigure[Batch size is 4]{
\label{fig_stl_b}
\includegraphics[width=0.4\columnwidth]{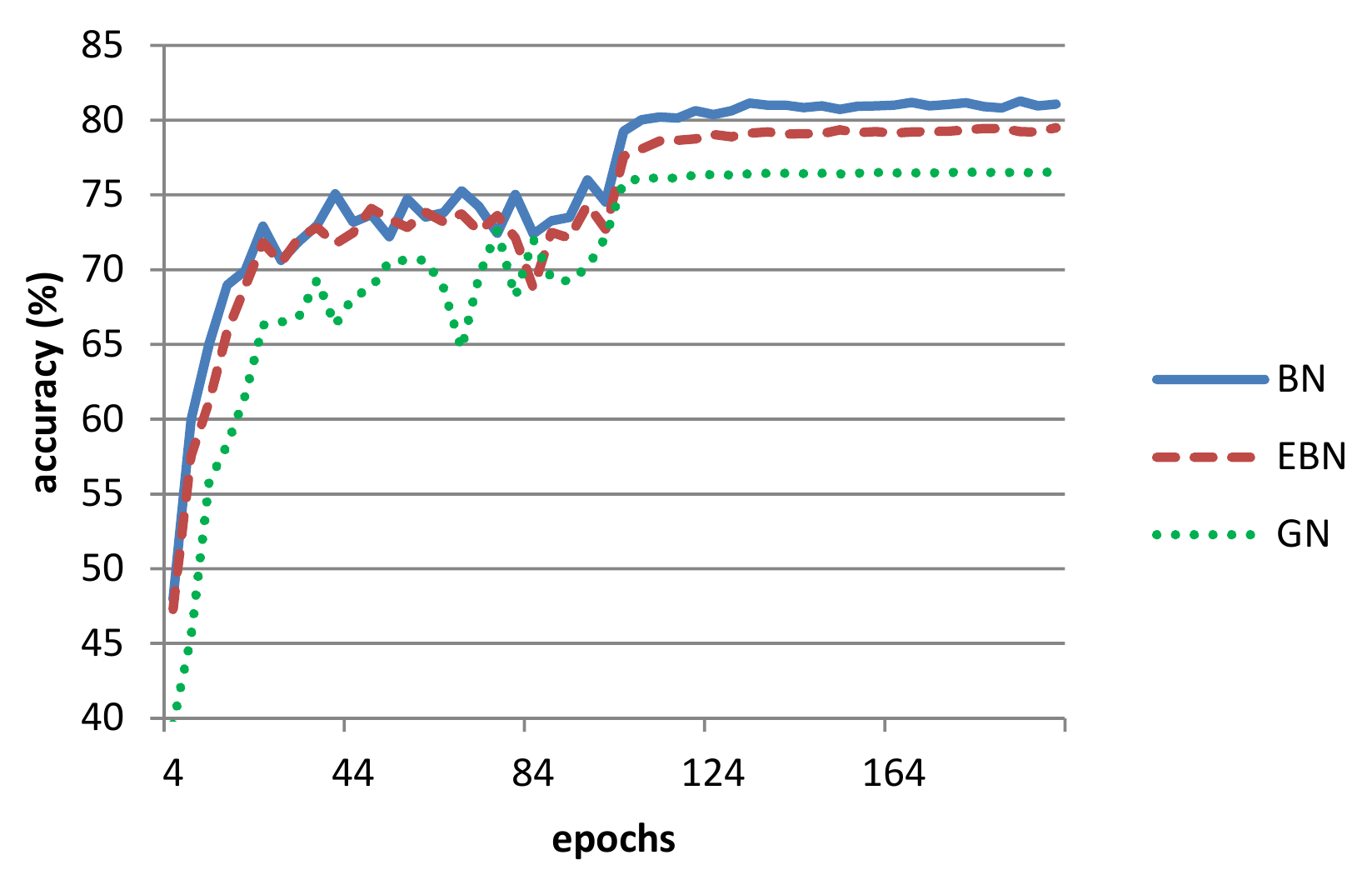}}
\centering
\subfigure[Batch size is 2]{
\label{fig_stl_b}
\includegraphics[width=0.4\columnwidth]{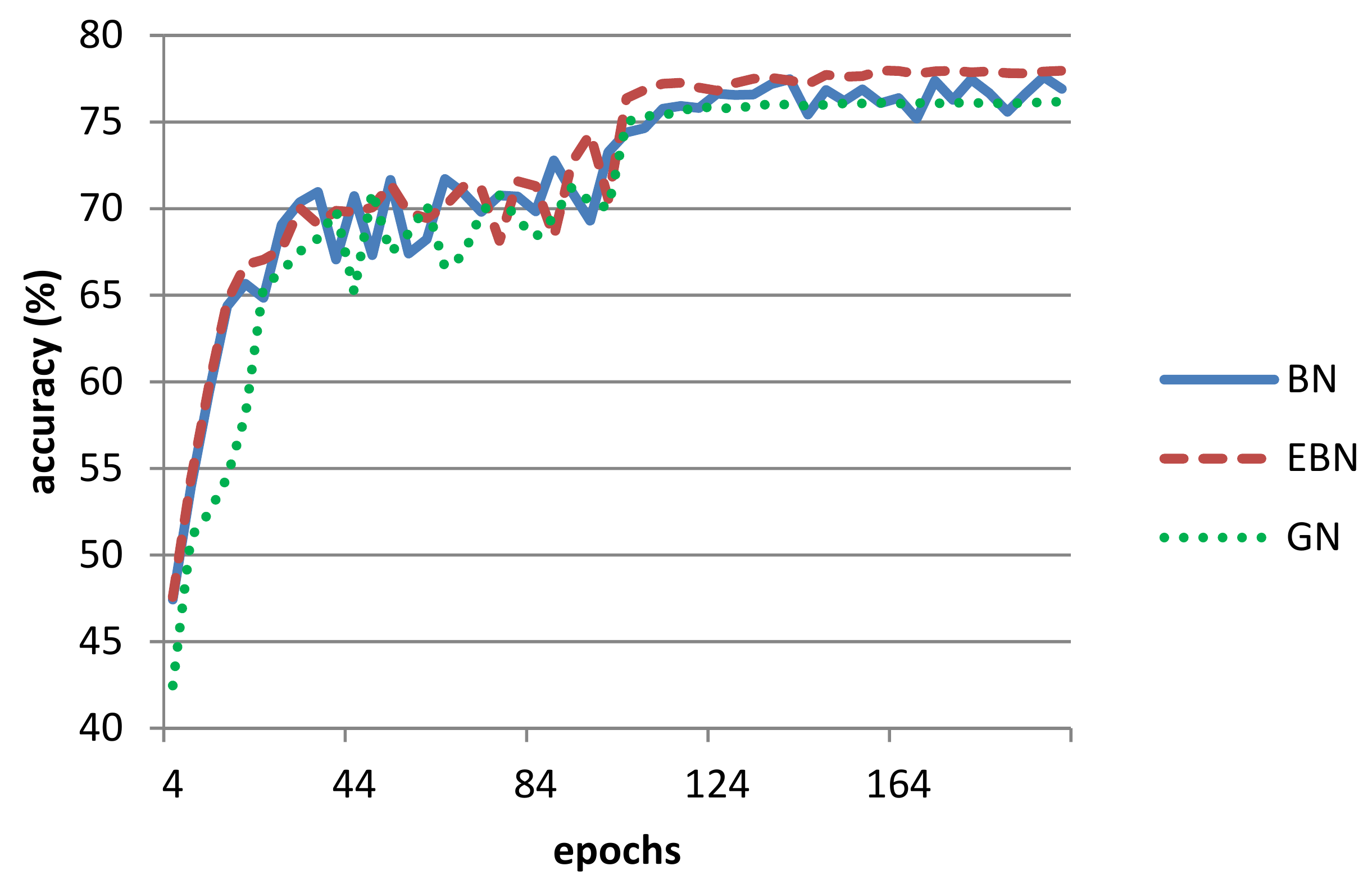}}
\caption{The test accuracy of ResNet18 on STL vs. the number of training epoch, with different normalization methods}
\label{fig_stl}
\end{figure}

The results of STL-10 are shown in Fig \ref{fig_stl}  and Table \ref{tab_stl}. With the batch size of 128 and 4, BN achieves best performance, and EBN performs better than GN. With the batch size of 4, all normalizations achieve better results than with the batch size of 128. STL-10 has fewer labeled training examples in total, thus the batch size could be smaller when training. That is to say, the batch size of 4 has not triggered the problem of small batch size of batch normalization. When using the batch size of 2, the performance of BN decreases to 76.91\%, and EBN achieves highest accuracy 77.96\%.

\begin{table}[!htb]
\caption{The test accuracy (\%) of ResNet18 on STL}
\label{tab_stl}
\centering
\begin{tabular}{l|lll}
\hline
  & BN & EBN & GN \\
\hline
batch size=128     & 78.65  & 	75.57 & 72.66     \\
batch size=4       & 81.04  & 79.3 & 76.49     \\
batch size=2       & 76.91  & 77.96 & 76.17     \\
\hline
\end{tabular}
\end{table}

\subsubsection{Summary}  The STL-10 dataset has higher resolution and fewer total training examples comparing to CIFAR dataset. Relatively, the batch size of 4 is not small enough to deteriorate the effect of BN.
When using smaller batch size of 2, EBN achieves the best performance. 

\subsection{ImageNet}

ImageNet classification dataset \cite{russakovsky2015imagenet} has 1.28M training images and 50,000 validation images with 1000 classes.  To augment data, the training images are cropped with random size of 0.08 to 1.0 of the original size and a random aspect ratio of 3/4 to 4/3 of the original aspect ratio,  and then resized to 224x224. Then random horizontal flipping is made. The validation image is resized to 256x256,  and then cropped by 224x224 at the center. Each channel of the input is normalized into 0 mean and 1 std globally.
Weight decay of 0.0001, and SGD with 0.9 momentum are used. 

On ImageNet, we evaluate the ResNet18 with BN, EBN and GN. We use 4 GPUs. We evaluate batch sizes of 64 and 4 images per GPU.   The mean and the standard deviation of BN and EBN are computed within each GPU. For the batch size of 64, the initial learning rate is set to 0.1. And for the batch size of 4, the initial learning rate is set to $0.1*4/64$, following the linear scaling rule. We train the network for 100 epochs, and decrease the learning rate by 10x at 30, 60 and 90 epochs.

\begin{figure}[!htb]
\centering
\subfigure[Batch size is 64 per GPU]{
\label{fig_imagenet_a}
\includegraphics[width=0.4\columnwidth]{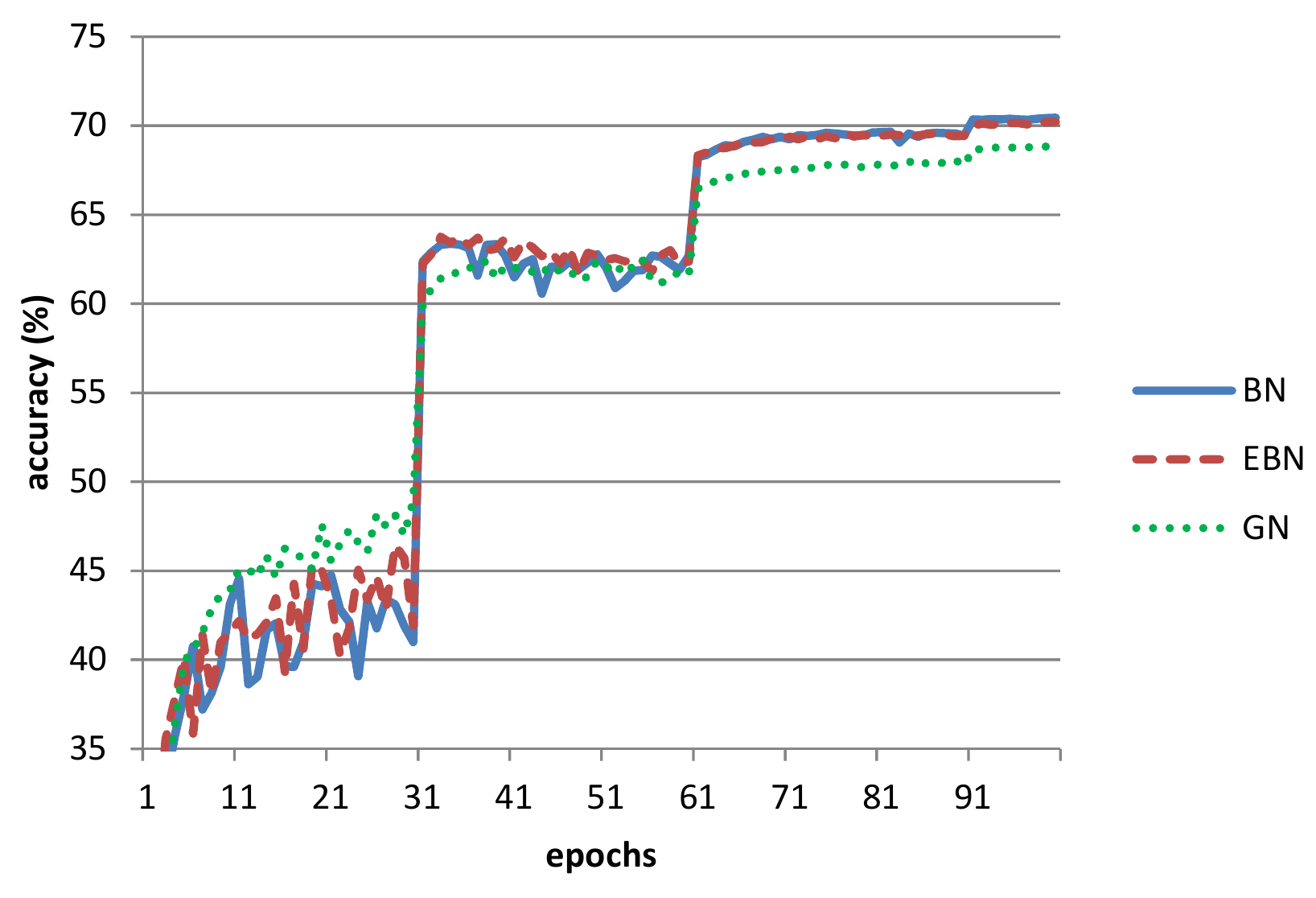}}
\centering
\subfigure[Batch size is 4 per GPU]{
\label{fig_imagenet_b}
\includegraphics[width=0.4\columnwidth]{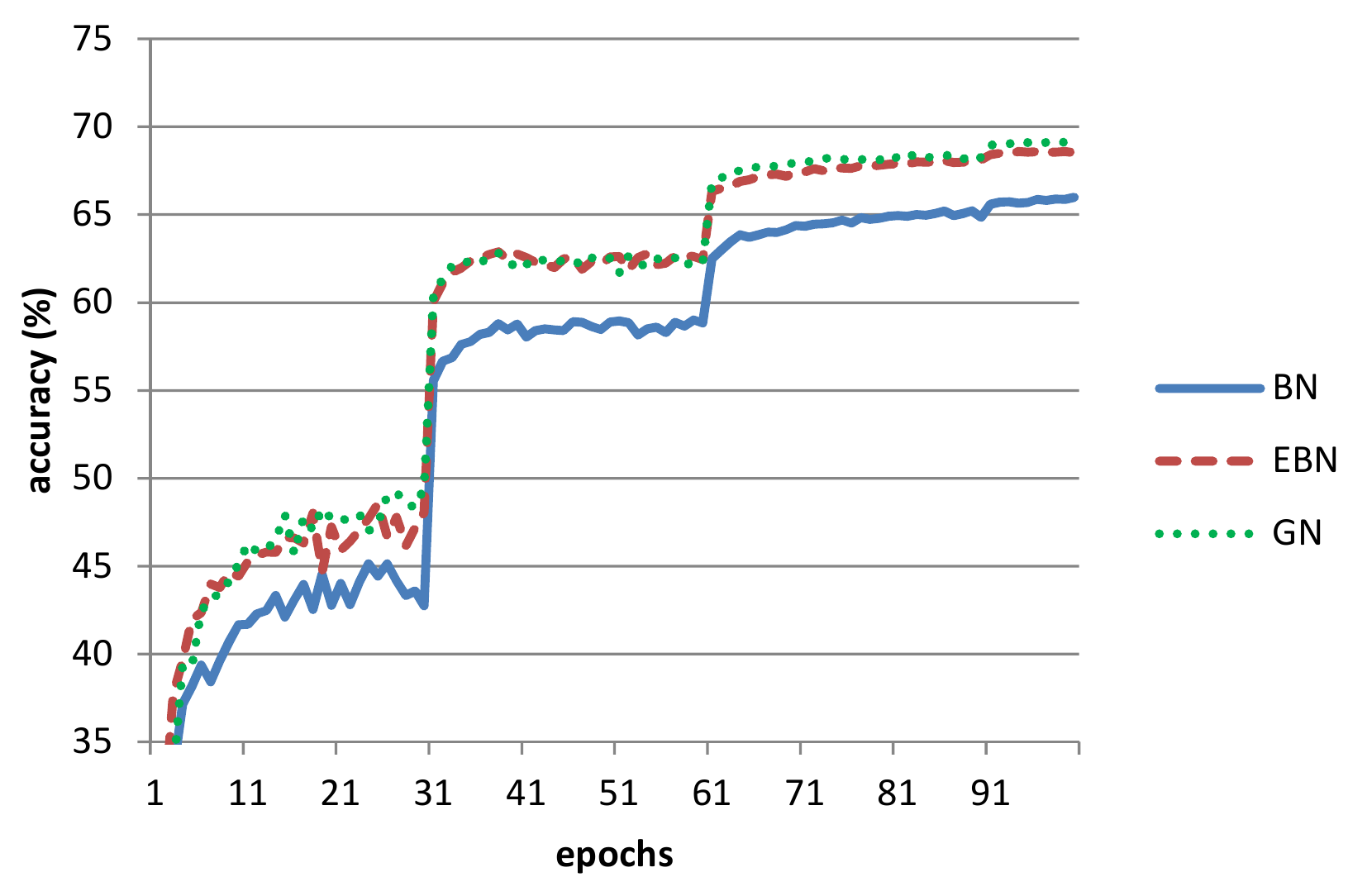}}
\caption{The test accuracy of ResNet18 on ImageNet vs. the number of training epoch, with different normalization methods}
\label{fig_imagenet}
\end{figure}

The results of ImageNet are shown in Fig \ref{fig_imagenet}  and Table \ref{tab_imagenet}.  From Fig \ref{fig_imagenet_a}, we can see EBN achieves close performance to BN, better than GN, in the case of large batch size. From Fig \ref{fig_imagenet_b}, we can see EBN achieves close performance to GN, better than BN, in the case of small batch size. As shown in Table \ref{tab_imagenet}, 
with the batch size of 64 per GPU, BN achieves the best validation accuracy of 70.37\%. EBN has 70.12\%, worse than BN by 0.25\%, but better than GN by 1.35\%.  With the batch size of 4 per GPU, the accuracy of BN decreases to 65.78\%.  GN achieves the best accuracy of 69.08\%. EBN achieves 68.54\%, worse than GN by 0.54\%, better than BN by 2.76\%.

\begin{table}[!ht]
\caption{The test accuracy (\%) of ResNet18 on ImageNet}
\label{tab_imagenet}
\centering
\begin{tabular}{l|lll}
\hline
  & BN & EBN & GN \\
\hline
batch size=128     & 70.37  & 	70.12 & 68.77     \\
batch size=4       & 65.78  & 68.54 & 69.08     \\
\hline
\end{tabular}
\end{table}

\subsubsection{Summary}  In the case of large batch size, EBN achieves close performance to BN, better than GN.  In the case of small batch size, EBN achieves close performance to GN, better than BN.

\section{Conclusion}

In this paper, we propose a simple but effective method, called extended batch normalization. The key difference from other normalization methods is that extended batch normalization computes the mean and the standard deviation from different set of pixels. 
To maintain the advantage of batch normalization, extended batch normalization computes the mean along the (N, H, W) dimensions just as the standard batch normalization. To alleviate the problems caused by small batch size, extended batch normalization enlarges the pixel set from which the standard deviation is computed along (N, C, H, W) dimensions. 
The experiments show that extended batch normalization alleviates the problem of batch normalization with small batch size while achieving close performances to batch normalization with large batch size.

Many other techniques could be equipped with extended batch normalization, e.g. kalman normalization \cite{wang2018kalman}, switchable normalization\cite{luo2018differentiable}.  
Moreover, as the same as batch normalization, the moving average of minibatch statistics is maintained during training and used during evaluation in extended batch normalization.  
As pointed by \cite{ioffe2017batch} and \cite{singh2019evalnorm}, that inconsistence hurts the performance of batch normalization in the case of small batch size. We will investigate the inconsistence between the training and the evaluation in extended batch normalization.

\clearpage
%
%
\bibliographystyle{splncs04}
\bibliography{extended_bn}

\begin{thebibliography}{10}
\providecommand{\url}[1]{\texttt{#1}}
\providecommand{\urlprefix}{URL }
\providecommand{\doi}[1]{https://doi.org/#1}

\bibitem{arpit2016normalization}
Arpit, D., Zhou, Y., Kota, B.U., Govindaraju, V.: Normalization propagation: A
  parametric technique for removing internal covariate shift in deep networks.
  arXiv preprint arXiv:1603.01431  (2016)

\bibitem{ba2016layer}
Ba, J.L., Kiros, J.R., Hinton, G.E.: Layer normalization. arXiv preprint
  arXiv:1607.06450  (2016)

\bibitem{bjorck2018understanding}
Bjorck, N., Gomes, C.P., Selman, B., Weinberger, K.Q.: Understanding batch
  normalization. In: Advances in Neural Information Processing Systems. pp.
  7694--7705 (2018)

\bibitem{cooijmans2016recurrent}
Cooijmans, T., Ballas, N., Laurent, C., G{\"u}l{\c{c}}ehre, {\c{C}}.,
  Courville, A.: Recurrent batch normalization. arXiv preprint arXiv:1603.09025
   (2016)

\bibitem{goyal2017accurate}
Goyal, P., Doll{\'a}r, P., Girshick, R., Noordhuis, P., Wesolowski, L., Kyrola,
  A., Tulloch, A., Jia, Y., He, K.: Accurate, large minibatch sgd: Training
  imagenet in 1 hour. arXiv preprint arXiv:1706.02677  (2017)

\bibitem{he2016deep}
He, K., Zhang, X., Ren, S., Sun, J.: Deep residual learning for image
  recognition. In: Proceedings of the IEEE conference on computer vision and
  pattern recognition. pp. 770--778 (2016)

\bibitem{hoffer2018norm}
Hoffer, E., Banner, R., Golan, I., Soudry, D.: Norm matters: efficient and
  accurate normalization schemes in deep networks. In: Advances in Neural
  Information Processing Systems. pp. 2160--2170 (2018)

\bibitem{ioffe2017batch}
Ioffe, S.: Batch renormalization: Towards reducing minibatch dependence in
  batch-normalized models. In: Advances in neural information processing
  systems. pp. 1945--1953 (2017)

\bibitem{ioffe2015batch}
Ioffe, S., Szegedy, C.: Batch normalization: Accelerating deep network training
  by reducing internal covariate shift. In: Proceedings of The 32nd
  International Conference on Machine Learning. pp. 448--456 (2015)

\bibitem{klambauer2017self}
Klambauer, G., Unterthiner, T., Mayr, A., Hochreiter, S.: Self-normalizing
  neural networks. In: Advances in neural information processing systems. pp.
  971--980 (2017)

\bibitem{krizhevsky2009learning}
Krizhevsky, A., Hinton, G.: Learning multiple layers of features from tiny
  images  (2009)

\bibitem{lecun1998gradient}
LeCun, Y., Bottou, L., Bengio, Y., Haffner, P.: Gradient-based learning applied
  to document recognition. Proceedings of the IEEE  \textbf{86}(11),
  2278--2324 (1998)

\bibitem{lecun2012efficient}
LeCun, Y.A., Bottou, L., Orr, G.B., M{\"u}ller, K.R.: Efficient backprop. In:
  Neural networks: Tricks of the trade, pp. 9--48. Springer (2012)

\bibitem{luo2017cosine}
Luo, C., Zhan, J., Wang, L., Yang, Q.: Cosine normalization: Using cosine
  similarity instead of dot product in neural networks. arXiv preprint
  arXiv:1702.05870  (2017)

\bibitem{luo2018differentiable}
Luo, P., Ren, J., Peng, Z., Zhang, R., Li, J.: Differentiable
  learning-to-normalize via switchable normalization. arXiv preprint
  arXiv:1806.10779  (2018)

\bibitem{luo2018towards}
Luo, P., Wang, X., Shao, W., Peng, Z.: Towards understanding regularization in
  batch normalization  (2018)

\bibitem{miyato2018spectral}
Miyato, T., Kataoka, T., Koyama, M., Yoshida, Y.: Spectral normalization for
  generative adversarial networks. arXiv preprint arXiv:1802.05957  (2018)

\bibitem{nam2018batch}
Nam, H., Kim, H.E.: Batch-instance normalization for adaptively style-invariant
  neural networks. In: Advances in Neural Information Processing Systems. pp.
  2558--2567 (2018)

\bibitem{ren2016normalizing}
Ren, M., Liao, R., Urtasun, R., Sinz, F.H., Zemel, R.S.: Normalizing the
  normalizers: Comparing and extending network normalization schemes. arXiv
  preprint arXiv:1611.04520  (2016)

\bibitem{russakovsky2015imagenet}
Russakovsky, O., Deng, J., Su, H., Krause, J., Satheesh, S., Ma, S., Huang, Z.,
  Karpathy, A., Khosla, A., Bernstein, M., et~al.: Imagenet large scale visual
  recognition challenge. International journal of computer vision
  \textbf{115}(3),  211--252 (2015)

\bibitem{salimans2016improved}
Salimans, T., Goodfellow, I., Zaremba, W., Cheung, V., Radford, A., Chen, X.:
  Improved techniques for training gans. In: Advances in neural information
  processing systems. pp. 2234--2242 (2016)

\bibitem{salimans2016weight}
Salimans, T., Kingma, D.P.: Weight normalization: A simple reparameterization
  to accelerate training of deep neural networks. In: Advances in Neural
  Information Processing Systems. pp. 901--901 (2016)

\bibitem{santurkar2018does}
Santurkar, S., Tsipras, D., Ilyas, A., Madry, A.: How does batch normalization
  help optimization? In: Advances in Neural Information Processing Systems. pp.
  2483--2493 (2018)

\bibitem{simonyan2014very}
Simonyan, K., Zisserman, A.: Very deep convolutional networks for large-scale
  image recognition. arXiv preprint arXiv:1409.1556  (2014)

\bibitem{singh2019evalnorm}
Singh, S., Shrivastava, A.: Evalnorm: Estimating batch normalization statistics
  for evaluation. arXiv preprint arXiv:1904.06031  (2019)

\bibitem{srivastava2014dropout}
Srivastava, N., Hinton, G.E., Krizhevsky, A., Sutskever, I., Salakhutdinov, R.:
  Dropout: a simple way to prevent neural networks from overfitting. Journal of
  Machine Learning Research  \textbf{15}(1),  1929--1958 (2014)

\bibitem{ulyanov2016instance}
Ulyanov, D., Vedaldi, A., Lempitsky, V.: Instance normalization: The missing
  ingredient for fast stylization. arXiv preprint arXiv:1607.08022  (2016)

\bibitem{wang2018kalman}
Wang, G., Luo, P., Wang, X., Lin, L., et~al.: Kalman normalization: Normalizing
  internal representations across network layers. In: Advances in Neural
  Information Processing Systems. pp. 21--31 (2018)

\bibitem{wu2018group}
Wu, Y., He, K.: Group normalization. In: Proceedings of the European Conference
  on Computer Vision (ECCV). pp. 3--19 (2018)

\bibitem{yuan2019generalized}
Yuan, X., Feng, Z., Norton, M., Li, X.: Generalized batch normalization:
  Towards accelerating deep neural networks. In: Proceedings of the AAAI
  Conference on Artificial Intelligence. vol.~33, pp. 1682--1689 (2019)

\end{thebibliography}
\end{document}